\documentclass[letterpaper, 10 pt, conference]{ieeeconf}  

\IEEEoverridecommandlockouts                              

\usepackage{amsmath} 
\usepackage{amssymb}  
\usepackage{cite}                      
\usepackage[nolist]{acronym}
\usepackage{tikz}
\usepackage{adjustbox}
\usepackage[caption=false]{subfig}
\usepackage{tikzscale}
\usepackage{pgfplots}
\usepackage{svg}
\usepackage{xcolor}
\newcommand{\degree}{\ensuremath{^\circ}}

\definecolor{audigrey}{rgb}{0.2, 0.2, 0.2 }
\definecolor{semfence}{RGB}{50, 120, 255}
\definecolor{semmotorcycle}{RGB}{150, 60, 30}
\definecolor{semred}{RGB}{255, 0, 0}
\definecolor{semcar}{RGB}{245, 150, 100}
\definecolor{sembi}{RGB}{245, 230, 100}
\definecolor{semperson}{RGB}{30, 30, 255}
\definecolor{sempole}{RGB}{150, 240, 255}
\definecolor{semsidewalk}{RGB}{75, 0, 75}
\definecolor{semroad}{RGB}{255, 0, 255}
\definecolor{sembuilding}{RGB}{0,   200, 255}
\definecolor{semvegetation}{RGB}{0,   175,   0}

\newcommand*\rot{\rotatebox{90}}
\newcommand{\approach}{PyFu}

\title{\LARGE \bf
Deep Sensor Fusion with Pyramid Fusion Networks for \\ 3D Semantic Segmentation
}

\author{Hannah Schieber$^{*, 1}$, Fabian Duerr$^{*, 2}$, Torsten Schoen$^{3}$ and Jürgen Beyerer$^{2, 4}$
\thanks{$^*$Equal contribution}%
\thanks{$^{1}$Hannah Schieber is with Human-Centered Computing and Extended Reality, Friedrich-Alexander University (FAU) Erlangen-Nürnberg, Erlangen, Germany {\tt\small hannah.schieber@fau.de}}%
\thanks{$^{2}$Fabian Duerr and Jürgen Beyerer are with Vision and Fusion Laboratory, Karlsruhe Institute of Technology, Karlsruhe, Germany {\tt\small fabian.duerr@partner.kit.edu}}
\thanks{$^{3}$Torsten Schoen is with Research Institute AImotion Bavaria, Technische Hochschule Ingolstadt, Ingolstadt, Germany {\tt\small torsten.schoen@thi.de}}%
\thanks{$^{4}$Jürgen Beyerer is with Fraunhofer Institute of Optronics, System Technologies and Image Exploitation (IOSB), Fraunhofer Center of Machine Learning, Karlsruhe, Germany {\tt\small juergen.beyerer@iosb.fraunhofer.de}}%
}

\begin{document}

\begin{acronym}[Bspwwww.]  
\acro{api}[API]{Application Programming Interface}
\acro{aspp}[ASPP]{Atrous Spatial Pyramid Pooling}
\acroplural{ann}[ANN]{Artifical Neural Networks}
\acro{bev}[BEV]{Bird Eye View}
\acro{rbob}[BRB]{Bottleneck Residual Block}
\acroplural{rbob}[BRBs]{Bottleneck Residual Blocks}
\acro{cel}[CE-Loss]{Cross-Entropy-Loss}
\acro{cnn}[CNN]{Convolutional Neural Network}

\acro{crf}[CRF]{Conditional Random Fields}
\acro{das}[DAS]{Driving Assistance System}
\acroplural{das}[DAS]{Driving Assistance Systems}
\acro{dpc}[DPC]{Dense Prediction Cells}
\acro{dla}[DLA]{Deep Layer Aggregation}
\acro{dnn}[DNN]{Deep Neural Network}
\acroplural{dnn}[DNNs]{Deep Neural Networks}
\acro{fcn}[FCN]{Fully Convolutional Network}
\acroplural{fcn}[FCNs]{Fully Convolutional Networks}
\acro{fov}[FoV]{Field of View}
\acro{fv}[FV]{Front View}
\acro{fp}[FP]{False Positive}
\acro{fpn}[FPN]{Feature Pyramid Network}
\acro{fn}[FN]{False Negative}
\acro{hmi}[HMI]{Human-Machine-Interaction}
\acro{iabn}[iABN]{Inplace Activated Batch Normalization}
\acro{iot}[IoT]{Internet of Things}
\acro{iou}[IoU]{Intersection over Union}
\acro{irb}[IRB]{Inverted Residual Block}
\acroplural{irb}[IRBs]{Inverted Residual Blocks}
\acro{knn}[KNN]{k-Nearest-Neighbor}
\acro{LiDAR}[LiDAR]{Light Detection and Ranging}
\acro{lsfe}[LSFE]{Large Scale Feature Extractor}
\acro{mc}[MC]{Mismatch Correction Module}
\acro{miou}[mIoU]{mean Intersection over Union}
\acro{ml}[ML]{Machine Learning}
\acro{mlp}[MLP]{Multilayer Perception}
\acro{nn}[NN]{Neural Network}
\acroplural{nn}[NNs]{Neural Networks}
\acro{pfn}[PFN]{Pyramid Fusion Network}
\acro{ppm}[PPM]{Pyramid Pooling Module}
\acro{rv}[RV]{Range View}
\acro{roi}[ROI]{Region of Interest}
\acroplural{roi}[ROIs]{Region of Interests}
\acro{rbab}[BB]{Residual Basic Block}
\acroplural{rbab}[BBs]{Residual Basic Blocks}
\acro{spp}[SPP]{Spatial Pyramid Pooling}
\acro{sgd}[SGD]{Stochastic Gradient Descent}
\acro{sac}[SAC]{Spatially-Adaptive Convolutions}
\acro{tp}[TP]{True Positive}
\acro{tn}[TN]{True Negative}
\end{acronym}

\maketitle
\thispagestyle{empty}
\pagestyle{empty}

\begin{abstract}
Robust environment perception for autonomous vehicles is a tremendous challenge, which makes a diverse sensor set with e.g. camera, lidar and radar crucial. In the process of understanding the recorded sensor data, 3D semantic segmentation plays an important role. Therefore, this work presents a pyramid-based deep fusion architecture for lidar and camera to improve 3D semantic segmentation of traffic scenes. Individual sensor backbones extract feature maps of camera images and lidar point clouds. A novel \textit{Pyramid Fusion Backbone} fuses these feature maps at different scales and combines the multimodal features in a feature pyramid to compute valuable multimodal, multi-scale features. The \textit{Pyramid Fusion Head} aggregates these pyramid features and further refines them in a late fusion step, incorporating the final features of the sensor backbones. The approach is evaluated on two challenging outdoor datasets and different fusion strategies and setups are investigated. It outperforms recent range view based lidar approaches as well as all so far proposed fusion strategies and architectures.
\end{abstract}
%
%
\section{INTRODUCTION}
\begin{figure}[t!]
\fontsize{8pt}{8pt}\selectfont%
\def\svgwidth{\columnwidth}{
\begingroup%
  \makeatletter%
  \providecommand\color[2][]{%
    \errmessage{(Inkscape) Color is used for the text in Inkscape, but the package 'color.sty' is not loaded}%
    \renewcommand\color[2][]{}%
  }%
  \providecommand\transparent[1]{%
    \errmessage{(Inkscape) Transparency is used (non-zero) for the text in Inkscape, but the package 'transparent.sty' is not loaded}%
    \renewcommand\transparent[1]{}%
  }%
  \providecommand\rotatebox[2]{#2}%
  \newcommand*\fsize{\dimexpr\f@size pt\relax}%
  \newcommand*\lineheight[1]{\fontsize{\fsize}{#1\fsize}\selectfont}%
  \ifx\svgwidth\undefined%
    \setlength{\unitlength}{428.58265344bp}%
    \ifx\svgscale\undefined%
      \relax%
    \else%
      \setlength{\unitlength}{\unitlength * \real{\svgscale}}%
    \fi%
  \else%
    \setlength{\unitlength}{\svgwidth}%
  \fi%
  \global\let\svgwidth\undefined%
  \global\let\svgscale\undefined%
  \makeatother%
  \begin{picture}(1,0.88066225)%
    \lineheight{1}%
    \setlength\tabcolsep{0pt}%
    \put(0,0){\includegraphics[width=\unitlength,page=1]{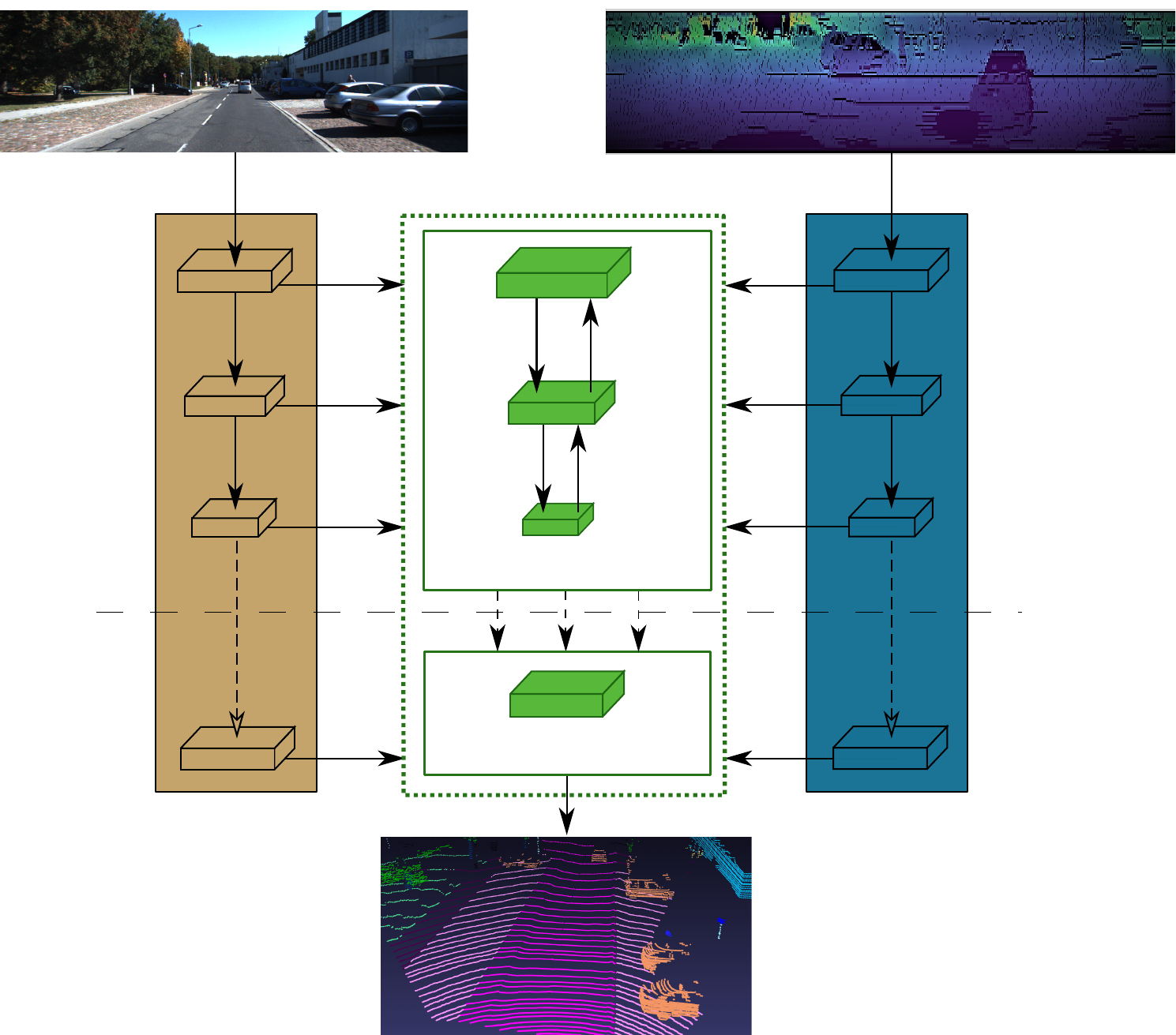}}%
    \put(0.199,0.17){\color[rgb]{0,0,0}\makebox(0,0)[t]{\smash{\begin{tabular}[t]{c}Camera \\Backbone\end{tabular}}}}%
    \put(0.32,0.715){\color[rgb]{0,0,0}\makebox(0,0)[lt]{\smash{\begin{tabular}[t]{l}Pyramid Fusion Network\end{tabular}}}}%
    \put(0.69,0.17){\color[rgb]{0,0,0}\makebox(0,0)[lt]{\smash{\begin{tabular}[t]{c}Lidar \\Backbone\end{tabular}}}}%
    \put(0.11,0.48){\color[rgb]{0,0,0}\rotatebox{90}{\makebox(0,0)[lt]{\smash{\begin{tabular}[t]{l}Encoder\end{tabular}}}}}%
    \put(0.11,0.23){\color[rgb]{0,0,0}\rotatebox{90}{\makebox(0,0)[lt]{\smash{\begin{tabular}[t]{l}Decoder\end{tabular}}}}}%
    \put(0.48076109,0.23716352){\color[rgb]{0,0,0}\makebox(0,0)[t]{\smash{\begin{tabular}[t]{c}\scriptsize{Pyramid Head}\end{tabular}}}}%
    \put(0.485,0.39){\color[rgb]{0,0,0}\makebox(0,0)[t]{\smash{\begin{tabular}[t]{c}\scriptsize{Pyramid Backbone}\end{tabular}}}}%
  \end{picture}%
\endgroup%
}
\caption{Modular deep fusion architecture with a novel Pyramid Fusion Network, which fuses camera and lidar features at multiple scales for improved 3D semantic segmentation.}
\label{fig:motivation}
\vspace{-0.1cm}
\end{figure}
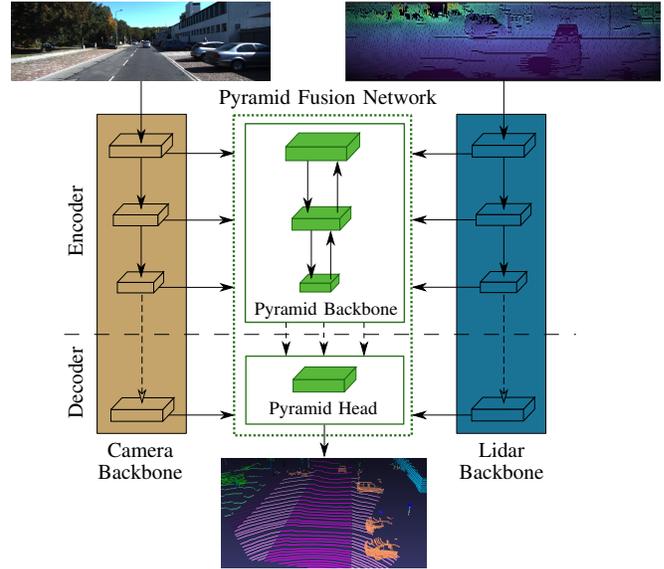
Semantic scene understanding plays a vital role in many robotic tasks. For a comprehensive understanding of complex 3D scenes it is crucial to exploit and combine different sensor modalities. A promising and complementary combination is the fusion of camera and lidar sensors. While cameras provide high resolution images but no geometric information, lidar point clouds provide valuable but sparse 3D geometric information, which gets even sparser with increasing distance. As a result, the fusion of camera images and 3D point clouds offers great potential. \\
One important aspect of 3D scene understanding is 3D semantic segmentation, which assigns a class label to every individual 3D point. While many approaches exist, which improve image semantic segmentation with depth information, the exploitation of camera information to improve 3D semantic segmentation is much less investigated. In general, various different representations for point clouds have been proposed, to allow the usage of a \ac{cnn}, e.g. projection-based~\cite{8967762, cortinhal2020salsanext} or point-based~\cite{qi2017pointnet, qi2017pointnetplusplus}. One promising projection-based representation is the range view, based on a spherical projection, as it allows an intuitive fusion with camera images. \\
Commonly used strategies~\cite{chen2017multiview} are the fusion of raw input data (early fusion), of feature maps (deep fusion) and the fusion of predictions (late fusion). While early and late fusion only fuse once and at one scale, deep fusion offers the possibility to fuse at multiple location and scales. Feature pyramids are frequently used in image processing to recognize multi-scale content. Therefore, they are a great starting point for multi-scale deep feature fusion.  \\
Motivated by these findings, we present a novel pyramid-based deep fusion approach, see Fig.~\ref{fig:motivation}, to exploit multi-scale fusion of lidar and camera for improved 3D semantic segmentation. Lidar features in range view space are fused with transformed camera features at different scales inside a feature pyramid by a novel \textit{Pyramid Fusion Backbone}. The proposed \textit{Pyramid Fusion Head} aggregates the multimodal, multi-scale features and refines them in a late fusion step. The overall \textit{Pyramid Fusion Network} improves the results significantly. To summarize, our contributions are:
\begin{itemize}
    \item A modular multi-scale deep fusion architecture, consisting of exchangeable sensor backbones and a novel Pyramid Fusion Network.
    \item A Pyramid Fusion Backbone for multi-scale feature fusion of lidar and camera in range view space.
    \item A Pyramid Fusion Head for aggregation and refinement of multimodel, multi-scale pyramid features.
\end{itemize}
%
\section{RELATED WORK}

\subsection{2D Semantic Segmentation}
A pioneer network architecture for the task of semantic segmentation were \acp{fcn}~\cite{long2015fully}. Fully convolutional architectures are designed for end-to-end pixel-level predictions as they replace the fully connected layers from \acsp{cnn} with convolutions. Since the original \ac{fcn} struggles to capture the global context of a scene~\cite{Zhao_2017_CVPR}, new architectures arised~\cite{mohan2021efficientps,Zhao_2017_CVPR, wang2020deep}, building upon pyramid features for multi-scale context aggregation, gathering the global context while preserving fine details. \\ PSPNet~\cite{Zhao_2017_CVPR} applies a \ac{ppm} that combines different scales of the last feature maps. As a result, the network is able to capture context as well as fine details of a scene. Further approaches like HRNetV2~\cite{wang2020deep} exploit pyramid features already in the backbone for feature extraction. For the related task of panoptic segmentation, EfficientPS~\cite{mohan2021efficientps} combines features at various scales bottom-up as well top-down by applying a two-way \ac{fpn}~\cite{lin2017feature}. Afterwards, a semantic head is applied, containing the modules \ac{lsfe}, \ac{dpc}~\cite{chen2018searching} and \ac{mc} to capture large and small scale features for semantic segmentation. 
\subsection{3D Semantic Segmentation}
In contrast to applying \acsp{cnn} on image data arranged in regular grids, they cannot be applied to 3D point clouds directly. Therefore, several representations and specialized architectures have been developed. \\
The pioneer approach directly working on the unstructured raw data is PointNet~\cite{qi2017pointnet}, which applies a shared multilayer perceptron to extract features for every input point. As it has to be invariant to any input permutations, a symmetric operation is used to aggregate features. Its successor PointNet++~\cite{qi2017pointnetplusplus} exploits spatial relationships between features by a recursive hierarchical grouping of the points. \\
Approaches, which do not process the raw point clouds transform them into a discrete space, like 2D or 3D grid. One efficient and promising 2D grid representation, based on a spherical projection, is the so-called range view. SqueezeSeg~\cite{wu2017squeezeseg} was one of the first approaches utilizing this representation for road object segmentation. The latest approach, SqueezeSegV3~\cite{xu2020squeezesegv3} uses Spatially-Adaptive Convolutions to counteract the varying feature distribution of the range view. RangeNet++~\cite{8967762} proposes an efficient kNN-based post-processing step to overcome some of the drawbacks induced by the spherical projection. Compared to previous approaches, SalsaNext~\cite{cortinhal2020salsanext} improves individual aspects of the network architecture, e.g. pixel-shuffle layer for decoding and the usage of Lov\'asz-Softmax-Loss~\cite{berman2018lovaszsoftmax}. Another adaption of convolutions is used in~\cite{razani2021litehdseg}. This approach applies light-weight harmonic dense convolutions to process the range view in real-time and shows promising results.
Furthermore, hybrid methods exploiting multiple representations emerged~\cite{tang2020searching,xu2021rpvnet}. 
\subsection{3D Multi-Sensor Fusion}
Multi-sensor fusion continuously gains attention for different tasks of computer vision. The combination of camera and lidar features is mainly tackled for 3D object detection. The dense fusion of features required for dense predictions like semantic segmentation is only investigated by a few works~\cite{liang2020multitask,liang2020deep, meyer2019sensor, fabian_wip}. \\ In~\cite{liang2020multitask}, a dense and region of interest based fusion is applied for multiple tasks, including 3D object detection. Another 3D object detection approach~\cite{liang2020deep} applies continuous convolutions to combine dense camera and lidar bird's eye view features. The continuous fusion layers fuse multi-scale image features with lidar feature maps at various scales in their network. \\
LaserNet++~\cite{meyer2019sensor} was proposed for the task of object detection and semantic segmentation. It first processes the camera images by a residual network. Applying a projection mapping, the camera features are transformed into the range view. Afterwards, the concatenated feature maps are fed into LaserNet~\cite{meyer2019lasernet}. Fusion3DSeg~\cite{fabian_wip} applies an iterative fusion strategy for camera and lidar features. Within Fusion3DSeg, camera and range view features are fused following an iterative deep aggregation strategy for an iterative multi-scale feature fusion. The final features are further combined with point-based features from a 3D branch, instead of the commonly used kNN-based post-processing~\cite{8967762}. \\
In contrast to~\cite{liang2020multitask,liang2020deep} the presented approach is modular and the individual sensor backbones are independent from each other, because no image features are fed to the lidar backbone. Additionally, a novel two-way pyramid fusion strategy is proposed. LaserNet++~\cite{meyer2019lasernet} on the other side only fuses once and does not apply multi-scale fusion. Fusion3DSeg~\cite{fabian_wip} is the most related work and applies an iterative fusion strategy, which considerably differs from our parallel bottom-up and top-down pyramid strategy.
%
%
\begin{figure*}[!t]
    \centering
    \subfloat[a) Overall architecture.]{%
		\fontsize{8pt}{8pt}\selectfont%
        \def\svgwidth{0.6\textwidth}{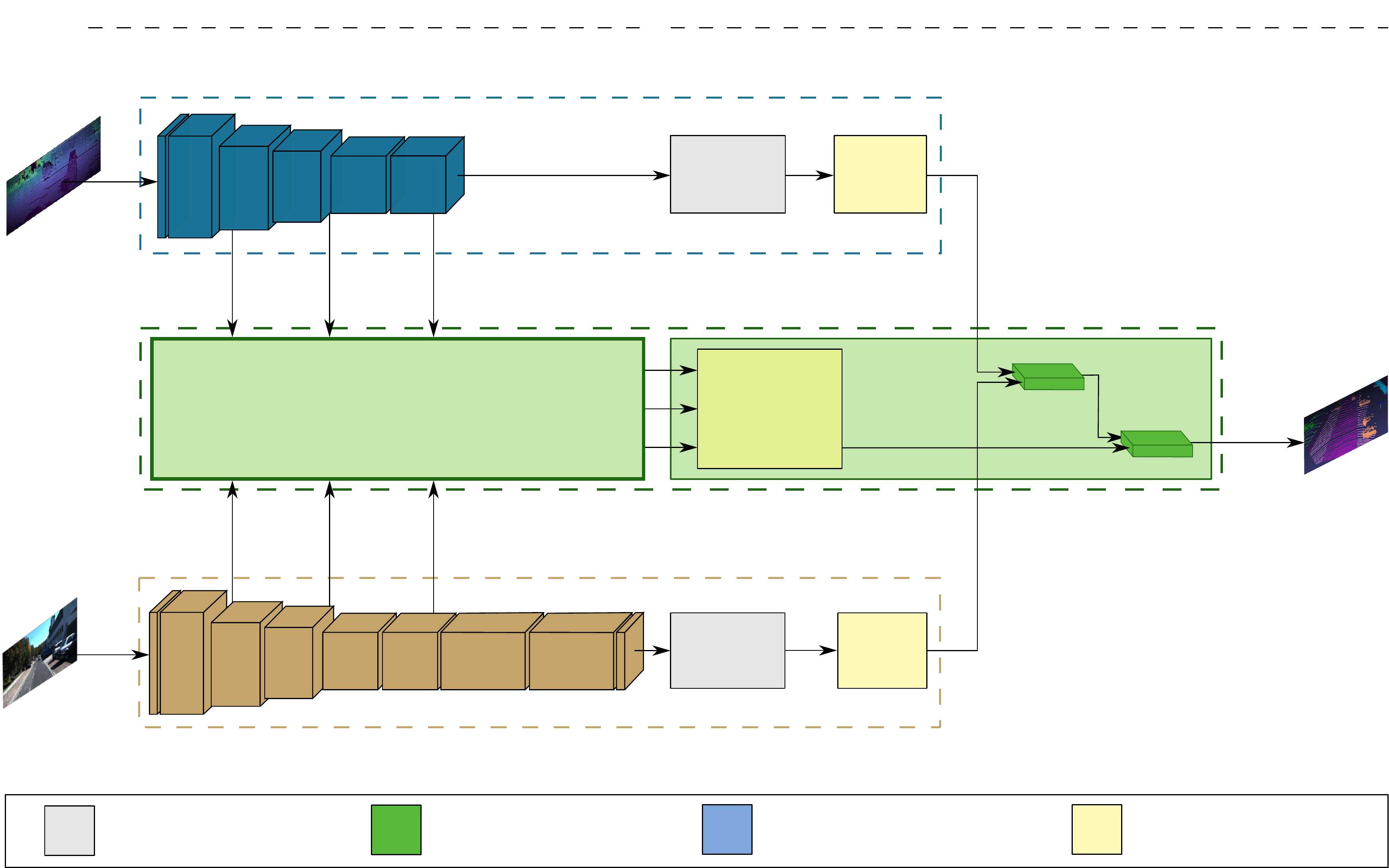}
        \label{fig:overall_arch}
    }%
    \hfill
    \subfloat[b) Pyramid Fusion Backbone]{
		\fontsize{8pt}{8pt}\selectfont%
        \def\svgwidth{0.35\textwidth}{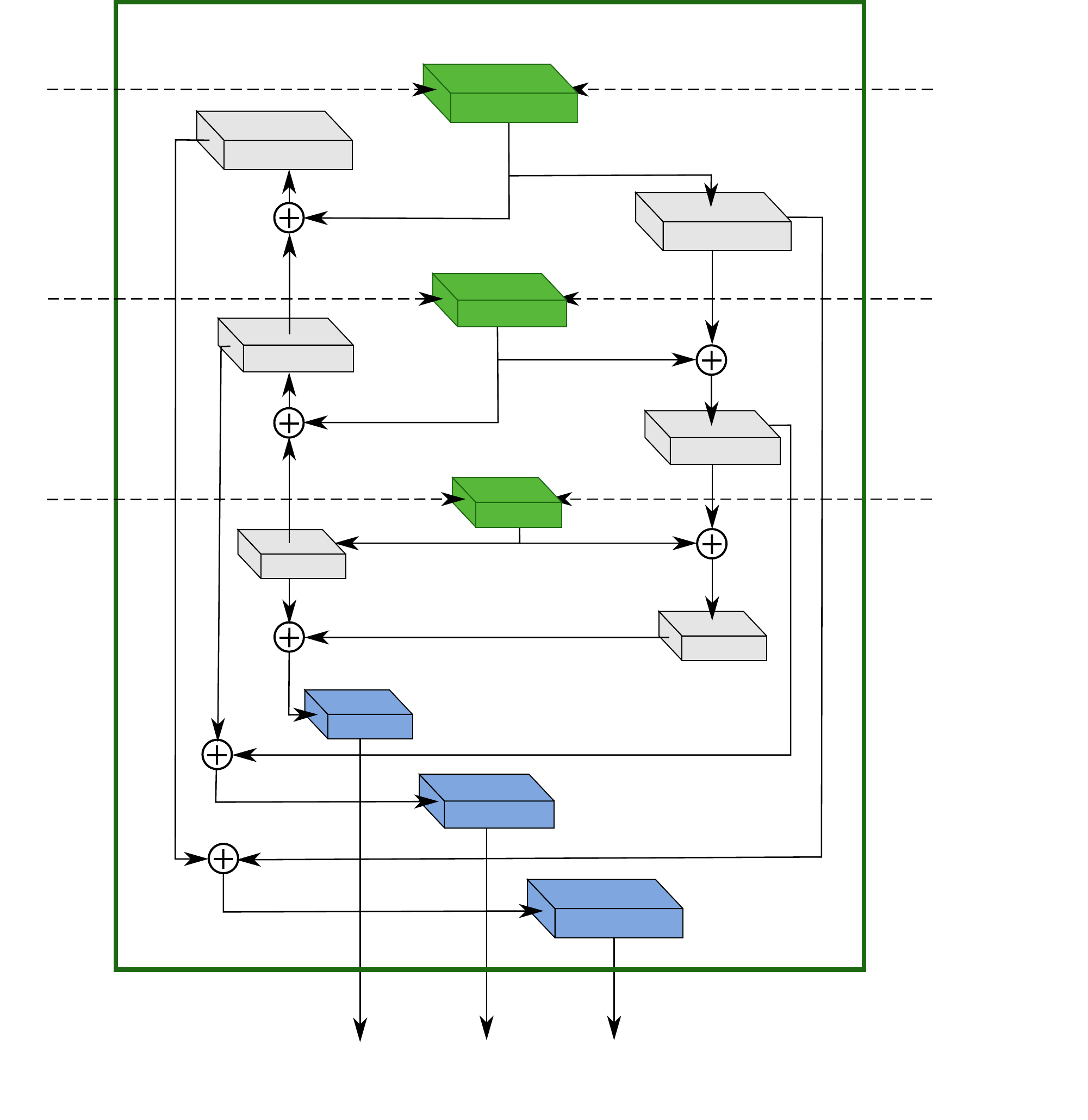}
        \label{fig:pfn_detail}
    }
    \caption{Overall architecture (a) with the Pyramid Fusion Network and both sensor backbones. The edges are labeled with the number of feature channels and the downsampling rate. Details of the Pyramid Fusion Backbone and its two-way pyramid structure is shown in (b).}
    \vspace{-0.3cm}
\end{figure*}
\section{Pyramid Fusion Networks for 3D Semantic Segmentation}
The proposed approach for deep sensor fusion \textit{\approach} consists of four main components, which are presented in the following section. Starting with lidar and camera backbone, which compute features for the individual sensor data, followed by a novel \textit{Pyramid Fusion Backbone} fusing the encoder features of both modalities at different scales in a top-down and bottom-up fashion. The \textit{Pyramid Fusion Head} combines these features and fuses them with the final output of both sensor backbone decoders in a late fusion step. The overall architecture is illustrated in Fig.~\ref{fig:overall_arch}. The modularity and chosen training strategy allows the approach to handle camera unavailability, switch backbone or sensor without affecting the other and to jointly predict camera and lidar semantic segmentation. Therefore, both backbones are pretrained on their sensor data and frozen during training of the overall fusion architecture. As a result, the backbones are still able to predict single sensor semantics, as fallback for camera unavailability or additional camera segmentation.
\subsection{Lidar Backbone}
The lidar backbone computes features for the input point clouds, which are represented in range view, based on the spherical projection of~\cite{duerr2021lidarbased, fabian_wip}. Its architecture is motivated by EfficientPS~\cite{mohan2021efficientps} and adapted to the range view. Because of the smaller resolution of range images compared to camera images, especially vertically, the downsampling step in the first two stages is only performed horizontally. Additionally, we use EfficientNet-B1~\cite{tan2020efficientnet} as encoder and remove the last three stages. As a result, the two-way \ac{fpn} has only three stages instead of four and the output channels are reduced to 128, because EfficientNet-B1 uses less feature channels than EfficientNet-B5. As depicted in Fig.~\ref{fig:overall_arch}, the computed feature maps of stages three, four and six are provided to the Pyramid Fusion Backbone for the fusion with camera features. As a result of the removed \ac{fpn} stage, the corresponding \ac{dpc} module~\cite{mohan2021efficientps} is removed from the semantic head too. The head provides its output features to the late fusion of the Pyramid Fusion Head.
\subsection{Camera Backbone}
The first backbone we investigate is again EfficientPS, but with the original Efficient-B5 as encoder. In contrast to the lidar backbone, EfficientPS is used nearly unchanged as camera backbone. Again, the output of stages three, four and six are provided to the Pyramid Fusion Backbone. For the late fusion step in the Pyramid Fusion Head, the output of the semantic head is used. \\
Additionally, PSPNet with underlying ResNet101~\cite{Yu_2017_CVPR} is chosen as another backbone. The three feature maps from layers conv3\textunderscore4, conv4\textunderscore23 and conv5\textunderscore3 of the ResNet101 are provided as input to the Pyramid Fusion Backbone. The output of the \ac{ppm} is provided to the late fusion step. 
\subsection{Pyramid Fusion Network}
\begin{figure}[t!]
    \centering
    \fontsize{8pt}{8pt}\selectfont%
     \def\svgwidth{\columnwidth}{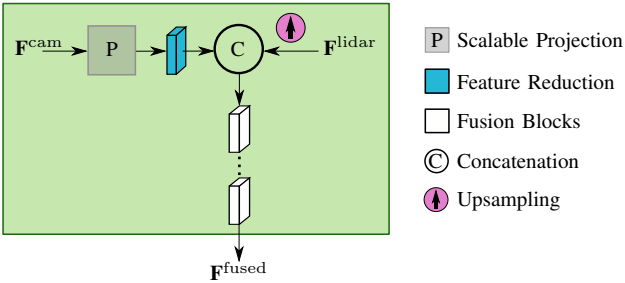}
    \caption{The fusion module transforms camera features into the lidar feature space and fuses them with lidar features.}
    \label{fig:fuseblocks}
    \vspace{-0.2cm}
\end{figure}

The key component of the fusion architecture is the Pyramid Fusion Network, which fuses lidar and camera features. A fusion module transforms the features into a common space, followed by a fusion step to combine both modalities. The Pyramid Fusion Backbone applies these modules at different scales and the resulting fused features are aggregated and combined in a top-down and bottom-up fashion, depicted in Fig.~\ref{fig:pfn_detail}. These multimodal, multi-scale features are combined and further refined in a late fusion step by the Pyramid Fusion Head. \vspace{0.075cm} \\ 
\textbf{Feature Transformation} 
To enable the fusion of lidar and camera, a common space is required. Therefore, a feature projection from camera image to range view space is needed. Additionally, the projection must be applicable at different feature map scales. To solve this task, we apply the scalable projection from Fusion3DSeg~\cite{9287974, fabian_wip}. The general idea is to create a mapping from camera image to range view coordinates based on the 3D points of the point cloud. Every point can be projected into the range view as well as into the camera image, creating the required link between camera image and range view coordinates. \vspace{0.075cm} \\
\textbf{Fusion Module} The feature transformation and fusion is performed by a fusion module, illustrated in Fig.~\ref{fig:fuseblocks}. First, both sensor feature maps are cropped to the overlapping field of view, as fusion is only possible in this area. Camera features are spatially transformed into the range view space by the mentioned feature transformation, followed by a learned feature projection to align lidar and camera feature space, implemented by one \ac{irb}~\cite{mohan2021efficientps}. Lidar features are bilinearly upsampled, because their feature map resolution is smaller compared to the camera, which allows to fuse distinct features from both sensors. The aligned lidar and camera features in range view space are then concatenated, followed by one or more residual blocks for the learned fusion. The module is designed to exploit different types and numbers of blocks to apply different fusion strategies. We investigate a bottleneck fusion strategy, based on a \ac{rbob} \cite{he2015deep}, and an inverted residual fusion strategy, using \acp{irb}. \vspace{0.075cm} \\ 
\textbf{Pyramid Fusion Backbone} The presented fusion module is incorporated into a two-way \ac{fpn} to fuse both modalities at different scales, followed by a bottom-up and top-down aggregation to compute multimodal, multi-scale features. From the lidar backbone, feature maps of three different scales, depicted in Fig.~\ref{fig:pfn_detail}, are passed to their corresponding fusion modules. There, the feature maps are upsampled to the targeted output resolution and fused with the feature maps from the camera backbone, which are also from three different scales. The three resulting feature maps are then aggregated in a bottom-up as well as top-down feature pyramid to compute multi-scale features. This way, the aggregation of multimodal features of different scale starts, on one side, with fine details and incorporates more and more context, and on the other side, with context and adds more and more details. Both pyramid outputs are finally combined and the resulting multimodal, multi-scale pyramid features are passed to the Pyramid Fusion Head. \vspace{0.075cm} \\
\textbf{Pyramid Fusion Head} The first step of the head is similar to the semantic head of the lidar backbone, which combines the three feature maps from the two-way \ac{fpn} and in this case from the Pyramid Fusion Backbone. While the latter fuses features from both encoders, an additional late fusion step, fuses the final features from lidar and camera decoder to further improve the features of the Pyramid Fusion Network. Therefore, the last feature maps from camera and lidar backbone are fused with the aggregated features of the Pyramid Fusion Backbone by two additional fusion modules, depicted in Fig.~\ref{fig:overall_arch}. \\
The final feature map is fed to a 1x1 convolution followed by a softmax activation function for the pixel-wise classification of the range view input. Applying the kNN-based post-processing~\cite{8967762} yields to the 3D semantic segmentation.
%
%
\section{EVALUATION}
The following evaluation is conducted on two challenging and large-scale outdoor datatsets, SemanticKITTI~\cite{Behley_2019_ICCV} and PandaSet~\cite{panda}, and in both cases the mean Intersection over-Union (mIoU) of the overlapping field of view of both sensors is reported to investigate our approach and compare it to other state-of-the-art approaches. \\
\textbf{SemanticKITTI} is a point-wise annotated dataset and based on 360\degree{} lidar scans of a Velodyne-HDL64E from the odometry task of the KITTI Vision Benchmark~\cite{Geiger2012CVPR}. It includes approximately 43,000 scans in 22 annotated sequences with 19 classes. From theses sequences, only the first eleven have publicly available ground truth. We report our results on the official validation sequence 08. Because no semantically labeled images for the odometry task exist, the data from the semantic segmentation challenge~\cite{Alhaija2018IJCV} is used to pretrain the camera backbone. It contains 200 annotated images and we use 0 - 149 for training and the remaining for validation. \\
\textbf{PandaSet}~\cite{panda} provides 6080 point-wise annotated lidar scans of a Pandar64, with corresponding camera images from the front center camera. For comparability with other approaches~\cite{duerr2021lidarbased} we group the labeled classes into a subset of 14 classes and follow the data split proposed in~\cite{duerr2021lidarbased}. %

\subsection{Implementation Details}
The presented approach is trained in mixed precision mode using distributed data parallel training on up to four Tesla V100 GPUs. During training, networks are optimized using weighted cross entropy loss, with weights \(w_c = \log\left(n_c / n \right)\), where \(n\) defines the total amount of points or pixel and \(n_c\) per class \(c\). Further, the poly learning rate scheduler is applied to decay the learning rate by \(1 - \left( i / i_{\text{max}}\right)^{0.9}\), where \(i\) denotes the iteration. If not stated otherwise, SGD optimizer with weight decay of 0.0001 is used. \\
\textbf{Lidar backbone} A batch size of 16 is used for both datasets and the initial learning rate is set to $0.07$ for SemanticKITTI and $0.001$ for PandaSet. The latter was optimized with Adam. To reduce overfitting, random horizontal flipping with a probability of 0.5 and random cropping, with a crop size of \(64 \times 1024\) are applied on both datasets. \\
\textbf{Camera Backbone} Both networks use pretrained weights from Cityscapes~\cite{cordts2016cityscapes} and are further trained with a batch size of four. For EfficientPS an initial learning rate of 0.0007 is applied,  while for PSPNet 0.0001 is used. Random horizontal flipping and gaussian blur with a probability of 0.5 are applied as well as random crops of size \(300 \times 600\) and random rotation with an angle uniformly drawn from \([-5\degree, 5\degree]\). If not stated otherwise EfficientPS is used as camera backbone. \\
\textbf{Deep Fusion} After both backbones have been pretrained and were frozen, the fusion approach on SemanticKITTI is trained with a batch size of 16 and a learning rate of 0.07. On PandaSet the batch size is set to 8, the initial learning rate to 0.001 and Adam optimizer is used. For data augmentation, random horizontal flipping is applied with a probability of 0.5. The resolution of the overlapping field of view in range view is \(45 \times 485\) and  \(61 \times 266\) respectively. Due to the small size of the sensor overlap no random cropping is applied. \\
Since no semantically labeled camera images for PandaSet exist, the camera backbone is not frozen for the corresponding experiments. To account for the pretraining, we set the learning rate to 0.0001 for EfficientPS and 0.001 for PSPNet.
\subsection{Pyramid Fusion Networks}
\begin{figure*}[t!]
\centering  
    \subfloat{
        \centering
        \adjustbox{raise=0.2cm}{\includegraphics[width=0.24\textwidth]{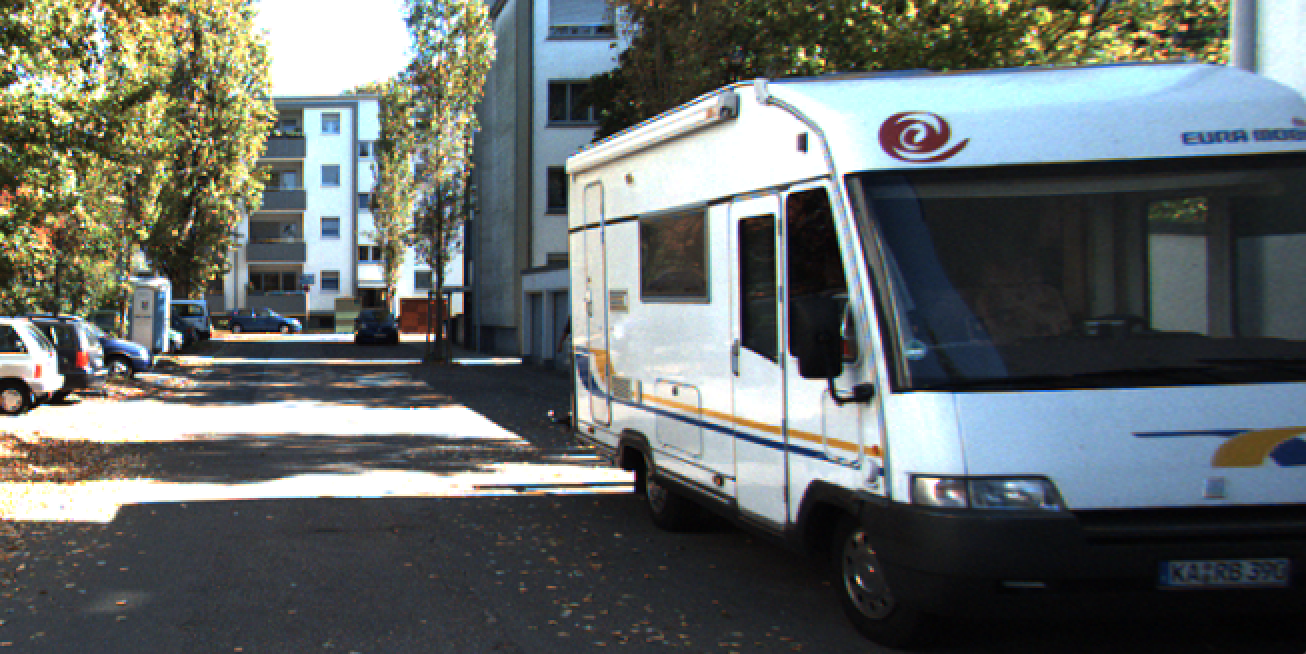}}
    }
    \subfloat{
        \centering
        \includegraphics[width=.235\textwidth]{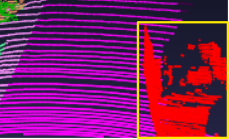}
    }
    \subfloat{
        \centering
        \includegraphics[width=.235\textwidth]{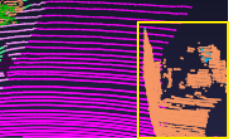}
    }
    \subfloat{
        \centering
        \includegraphics[width=.235\textwidth]{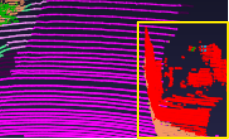}
    } \\
    \subfloat{
        \centering
        \adjustbox{raise=0.25cm}{\includegraphics[width=0.24\textwidth]{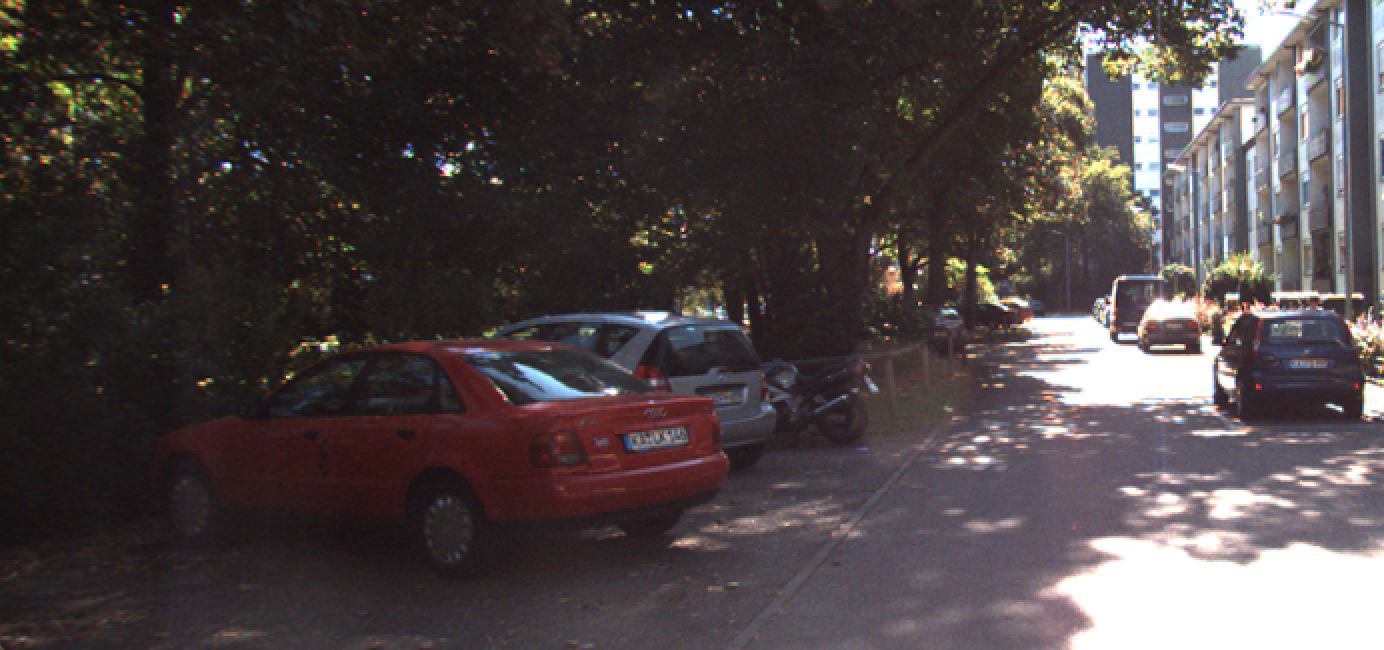}}
    }
    \subfloat{
        \centering
        \includegraphics[width=.235\textwidth]{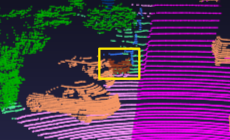}
    }
    \subfloat{
        \centering
        \includegraphics[width=.235\textwidth]{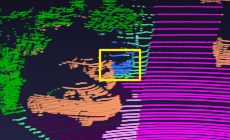}
    }
    \subfloat{
        \centering
        \includegraphics[width=.235\textwidth]{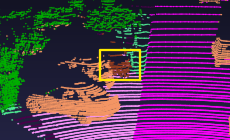}
    } \\
    \subfloat[(a) Camera]{
        \centering
        \includegraphics[width=0.24\textwidth]{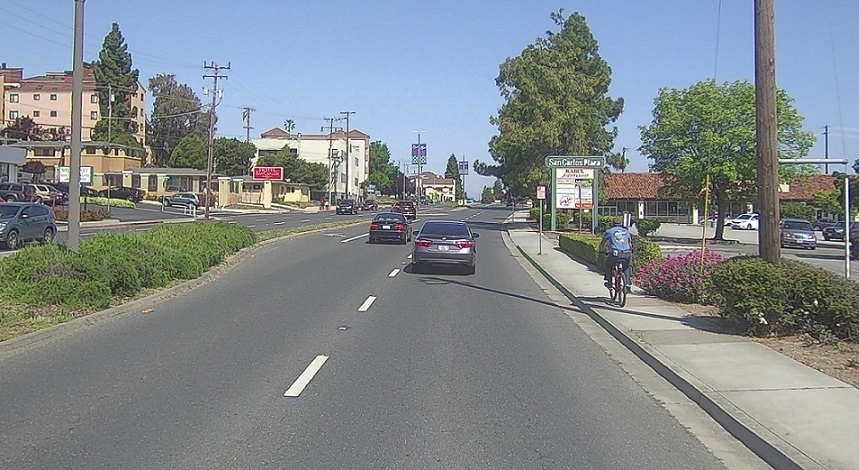}
    }
    \subfloat[(b) Ground Truth]{
        \centering
        \includegraphics[width=.235\textwidth]{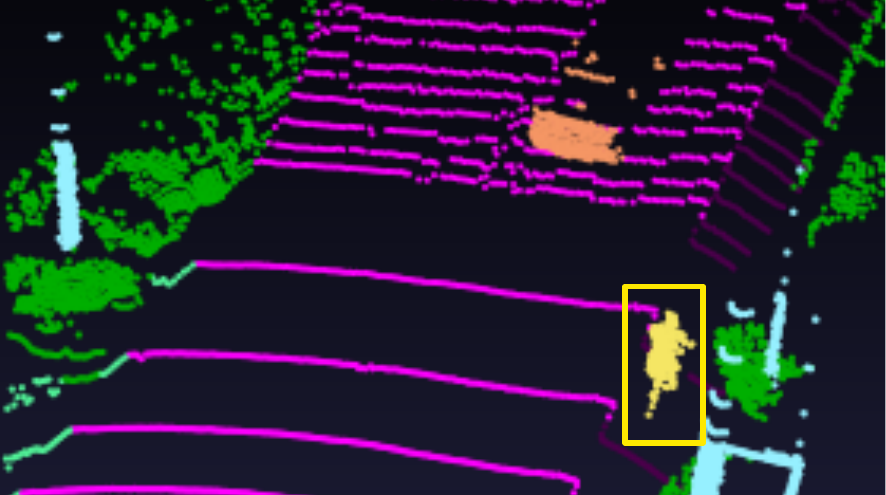}
    }
    \subfloat[(c) Baseline]{
        \centering
        \includegraphics[width=.235\textwidth]{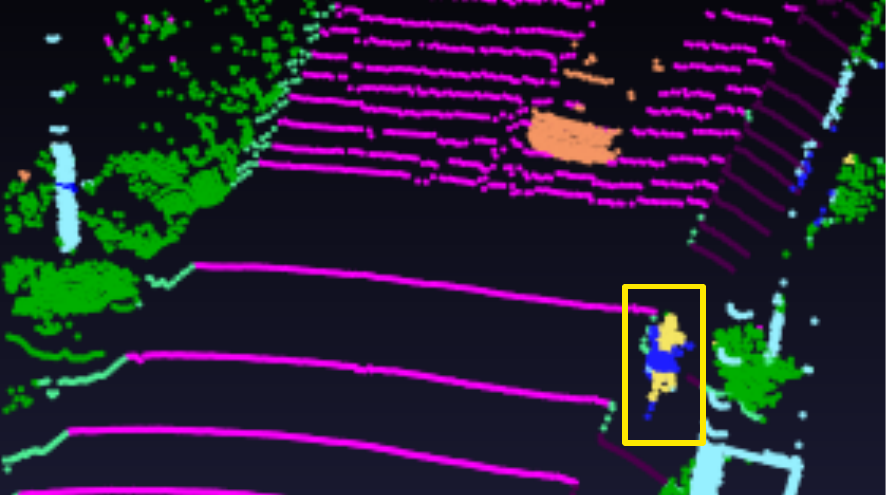}
    }
    \subfloat[(d) \approach]{
        \centering
        \includegraphics[width=.235\textwidth]{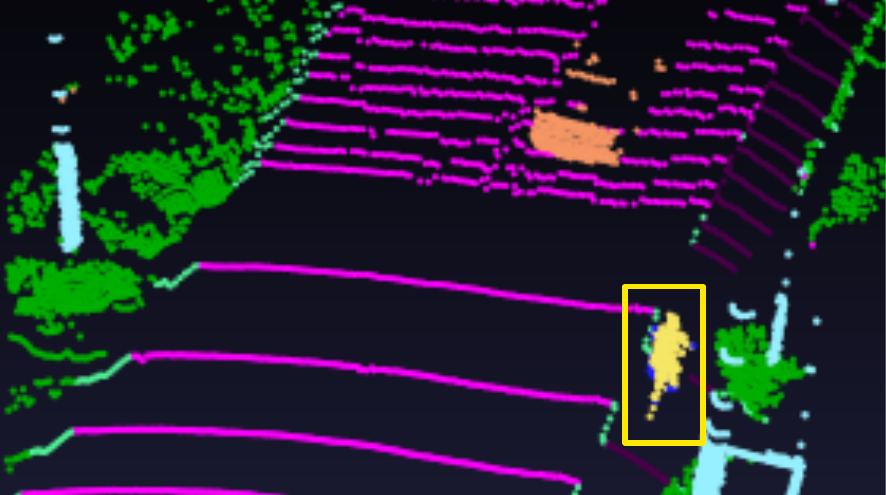}
    }
\caption[]{Qualitative results on SemanticKITTI and PandaSet. In the first scene of SemanticKITTI (top), the lidar backbone predicts the class car~\tikz \fill [semcar](0.1,0.1) rectangle (0.3,0.3); while \approach\text{} correctly predicts the campervan, which belongs to the class other-vehicle~\tikz \fill [semred](0.1,0.1) rectangle (0.3,0.3);. In the second scene of SemanticKITTI (center), the lidar backbone confuses the class fence~\tikz \fill [semfence](0.1,0.1) rectangle (0.3,0.3); and motorcycle~\tikz \fill [semmotorcycle] (0.1,0.1) rectangle (0.3,0.3);, while our fusion approach predicts the motorcycle correctly. In the third scene (bottom) and on PandaSet, \approach\text{} does not confuse parts of the bicyclist~\tikz \fill [sembi](0.1,0.1) rectangle (0.3,0.3); with person~\tikz \fill [semperson](0.1,0.1) rectangle (0.3,0.3);.}
\label{fig:qualitative_res}
\vspace{0.3cm}
\end{figure*}

\begin{table}[b]
\centering
\renewcommand{\arraystretch}{1.2}
\caption{Analysis of the fusion components late fusion (LF), Pyramid Fusion Backbone (PFB) and Pyramid Fusion Head (PFH) on the validation sequence of SemanticKITTI.}
\label{tab:abl_components}
\resizebox{0.8\columnwidth}{!}{
\begin{tabular}{c c c c | c} \hline 
    Backbone & LF & PFB & PFH  & \acs{miou} in \% \\ \hline
    \checkmark &            &            &            & 56.7 \\
    \checkmark & \checkmark &            &            & 57.9 \\ \hline
    \checkmark &            & \checkmark &            & 59.1 \\
    \checkmark &            & \checkmark & \checkmark & \textbf{60.6} \\  \hline
\end{tabular}}
\end{table}
The first experiment of our evaluation investigates the impact of the proposed approach and its components on SemanticKITTI. The results are denoted in Table~\ref{tab:abl_components}. Two important baselines are the lidar backbone as single sensor baseline and a late fusion strategy as fusion baseline, which fuses the final feature maps of both backbones with the presented fusion module. Adding our fusion backbone PFB improves the results significantly, which emphasises the value of the multimodal features, and additionally, emphasises the benefits of our multi-scale fusion strategy, which outperforms the late fusion. So far, the semantic head of the lidar backbone was used as pyramid head to aggregate the pyramid features. Deploying our fusion head PFH, which includes an additional late fusion step, further improves the results. Overall, \approach\text{} outperforms both baselines by $+3.9\%$ and $+2.7\%$ respectively, with an inference time of $48\,$ms. \\
In the next step, different fusion strategies inside the fusion module are investigated, with the results shown in Table~\ref{tab:abl_fusion_module}. First, the influence of varying strategies is evaluated for the pyramid backbone PFB. The bottleneck fusion strategy using a \ac{rbob} followed by a \ac{rbab} \cite{he2015deep} outperforms the inverted strategy relying on \acp{irb}. This holds also for the entire Pyramid Fusion Network. \\
Another experiment is performed on PandaSet to investigate different camera backbones. Like previously mentioned, the camera backbone is trained with the overall architecture on PandaSet because no semantically labeled image data exists. While no joint lidar and camera segmentation is possible in this case, it however shows that our approach is also trainable without semantically labeled images. Table~\ref{tab:abl_camera_backbones} reveals that EfficientPS as camera backbone considerably outperforms PSPNet and achieves a remarkable improvement of $+8.8\%$ over the baseline. Nevertheless, the lidar baseline is significantly outperformed in both cases confirming that the presented architecture works well with different camera backbones. EfficientPS offers another advantage by requiring less memory, making it possible to process camera images in original resolution. Otherwise, an initial downsampling by a factor of two is performed, which reduces performance. On SemanticKITTI, PSPNet works best as backbone, with an mIoU of $61.9\%$. For the latter and on SemanticKITTI the LF step does not provide any improvements. \\
Finally, Fig.~\ref{fig:qualitative_res} shows qualitative results for both datasets in three different scenes.
\begin{table}[t]
\centering
\renewcommand{\arraystretch}{1.2}
\caption{Evaluation of different fusion strategies on the validation set of SemanticKTTI.}
\label{tab:abl_fusion_module}
\resizebox{0.71\columnwidth}{!}{
\begin{tabular}{l|c|c} \hline 
    Approach & Fusion Strategy & \acs{miou} in \% \\ \hline
    Lidar Backbone &                      & 56.7  \\ \hline
    PFB		    & 1x \acs{irb}            & 58.5 \\
    PFB         & 2x \acs{irb}            & 58.8 \\
    PFB         & \acs{rbob} + \acs{rbab} & 59.1 \\  \hline
    PFB + PFH   & 2x \acs{irb}            & 60.2 \\  
    PFB + PFH   & \acs{rbob} + \acs{rbab} & \textbf{60.6} \\  \hline 
\end{tabular}}
\end{table}
\begin{table}[t]
\centering
\renewcommand{\arraystretch}{1.2}
\caption{Results for different camera backbones on the validation sequences of PandaSet.} 
\label{tab:abl_camera_backbones}
\resizebox{\columnwidth}{!}{
\begin{tabular}{l|c|c|c} \hline
    Approach    & Camera Backbone  & Add. Down. & \acs{miou} in \% \\ \hline 
    Lidar Backbone &            &               & 59.0 \\ \hline 
   	PFB         & EfficientPS   &               & 63.1 \\  
   	PFB + PFH   & EfficientPS   &               & \textbf{67.8} \\   
    PFB + PFH   & EfficientPS   & \checkmark    & 65.3 \\  \hline  
    PFB         & PSPNet        & \checkmark    & 60.0 \\  
    PFB + PFH   & PSPNet        & \checkmark    & 64.4 \\  \hline
\end{tabular}}
\end{table}
\subsection{Quantitative Results} 
In the next step, our approach is further compared with state-of-the art range view based approaches to evaluate the benefits of camera and lidar fusion, with the results for SemanticKITTI shown in Table~\ref{tab:comp_sota_sem_kitti}. It outperforms all lidar approaches in general and also for the majority of the individual classes. It is worth mentioning that the dominating part of the improvements results from the fusion and not from an already superior baseline. This underlines again the value of camera features to improve 3D semantic segmentation. Next, the comparison to other deep fusion approaches investigates our fusion architecture and strategy, see Table~\ref{tab:comp_sota_sem_kitti}. The pyramid fusion strategy outperforms all other fusion approaches and the superior performance of \approach\text{} and Fusion3DSeg~\cite{fabian_wip} underlines the benefits of multi-scale sensor fusion. Compared to the latter, especially for the classes motorcycle, other-vehicle and fence significantly better results are achieved. The other approaches have been evaluated on the overlapping field of view with KNN-based post-processing in~\cite{fabian_wip}. Razani et al.~\cite{razani2021litehdseg} could not be evaluated because no code is available. \\
\begin{table*}[t]
\centering
\renewcommand{\arraystretch}{1.25}
\caption{Comparison of \approach\text{} with state-of-the-art range view and multi-modal fusion approaches on sequence 08 of SemanticKITTI. Values are given as \acs{iou} in \%.}
\label{tab:comp_sota_sem_kitti}
\resizebox{\textwidth}{!} {
\begin{tabular}{l|c|c|ccccccccccccccccccc} \hline
Approach  &  \rot{modalities} & \rot{\acs{miou}} & \rot{car} & \rot{bicycle} & \rot{motorcycle}  & \rot{truck} & \rot{other-vehicle \textit{ }} & \rot{person} & \rot{bicyclist} & \rot{motorcyclist} & \rot{road} & \rot{parking} & \rot{sidewalk} & \rot{other-ground \textit{ }} & \rot{building} & \rot{fence} & \rot{vegetation} & \rot{trunk} &\rot{terrain} & \rot{pole} & \rot{traffic-sign} \\ \hline
RangeNet++~\cite{8967762}  & li & 52.5 & 90.3 & 27.8 & 48.6  & 34.3  & 26.6 & 55.3 & 69.2 & 0.0 & 94.6 & 39.5 & 76.4 & 0.0 & 84.6 & 52.9 & 84.0 & 55.9 & 70.3 & 52.6 & 34.4 \\
SqueezeSegV3~\cite{xu2020squeezesegv3}  & li & 55.6 & 91.6 & 37.9 & 53.4  & 47.6  & 48.4 & 63.1 & 57.9 & 0.0 & 94.8 & 42.0 & 77.6  & 0.0  & 86.0 & 55.8  & 85.0 & 58.7  & 70.3 & 52.2 & 34.4 \\ 
SalsaNext~\cite{cortinhal2020salsanext}  & li & 61.4 & 91.5 & 51.8 & 50.7  & \textbf{86.9} & 54.9 & \textbf{76.6} & 81.5 & 0.0 & 95.3 & 43.7  & 76.0 & 0.0 & 86.0 & 55.5  & 84.5 & 66.1 & 66.5 & 58.7 & 40.3 \\
Backbone (Ours) & li & 56.7 & 90.4 & 37.8 & 50.4 & 73.0 & 32.1 & 59.2 & 68.2 & 0.0 & 95.1 & \textbf{49.6} & 79.1 & 0.0 & 82.4 & 56.8 & 84.8 & 59.8 & \textbf{74.3} & 49.4 & 35.4 \\ \hline 
LaserNet++~\cite{meyer2019sensor} & li+c & 56.2 & 92.7 & 37.6 & 41.1 & 55.8 & 50.0 & 61.4 & 60.8 & 0.0 & 95.0 & 34.8 & 77.9 & 0.0  & 86.4  & 55.0 & 86.3 & 65.1 & 71.5 & 59.7 & 36.6 \\
Fusion3DSeg~\cite{fabian_wip} & li+c & 61.8 & \textbf{94.3} & \textbf{52.1} & 48.4 & 75.6 & 56.9 & 72.1 & \textbf{82.6} & 0.0 & \textbf{96.0} & 45.0 & \textbf{79.2} & 0.0 & 86.9 & 56.6 & \textbf{86.7} & \textbf{67.3}  & 72.2 & \textbf{60.6} & \textbf{42.3} \\ \hline 
PyFu (Ours) & li+c & \textbf{61.9} & 93.3 & 47.0 & \textbf{55.7} & 73.6 & \textbf{62.2} & 72.9 & 76.9 & 0.0 & 95.3 & 48.3 & 78.2 & 0.0  & \textbf{87.4} & \textbf{63.9} & 86.3 & 65.2 & 71.7 & 59.4  & 38.8 \\ \hline  
\end{tabular}}%
\end{table*}%
Finally, \approach\text{} is compared to other fusion approaches on PandaSet. Once more, our approach performs best and considerably outperforms all other approaches by an even higher margin. Looking at the individual classes, especially for truck~($+22.4$\%), other-vehicle~($+10.9$\%) and road\_barriers~($+9.1$\%) \approach\text{} works significantly best. In these cases, also our lidar backbone performs better than the lidar backbone of Fusion3DSeg. However, it shows that the presented fusion strategy is capable of improving already strong classes even further. For the classes car, person, building and background our approach achieves the best results despite the better performance of Fusion3DSeg's lidar backbone. If our approach does not accomplish the best results for a class, the other backbone also works better than ours. However, the pyramid fusion strategy reduces the gaps significantly. Overall, this shows the great potential of our fusion strategy and architecture. \\
\begin{table*}[t]
\centering
\renewcommand{\arraystretch}{1.1}
\caption{Comparison of \approach\text{} with state-of-the-art multi-modal fusion approaches on PandaSet. Values are given as \acs{iou} in \%.}
\label{tab:comp_sota_panda}
\begin{tabular}{l|c|c|cccccccccccccc} \hline
Approach & \rot{modalities} & \rot{\acs{miou}} & \rot{car}  & \rot{bicycle} & \rot{motorcycle } & \rot{truck} & \rot{other-vehicle \textit{ }} & \rot{person} & \rot{road} & \rot{road\_barriers \textit{ }} & \rot{sidewalk} & \rot{building} & \rot{vegetation} & \rot{terrain} & \rot{background  \textit{ }} & \rot{traffic-sign} \\ \hline
Backbone (Ours) & li & 59.0 & 94.6 & 28.3 & 22.9 & 68.7 & 77.4 & 45.5 & 95.6   & 29.0 & 67.5 & 81.8 & 83.0 & 59.0 & 56.9 & 16.0 \\ 
Lidar3DSeg~\cite{fabian_wip} & li & 63.1 & 95.8 & 41.9 & 45.8 & 55.2 & 61.3 & 65.5 & 96.2 & 25.4 & 72.5 & 86.8 & 88.8 & \textbf{64.4} & 68.6 & 15.6 \\ \hline
LaserNet++~\cite{meyer2019sensor} & li+c  & 59.7  & 94.4 & 23.6 & 38.5 & 27.3 & 72.1 & 61.2 & 95.2 & 26.6 & 72.3 & 85.5 & 90.6 & 62.7 & 67.7 & 18.1 \\
Fusion3DSeg~\cite{fabian_wip} & li+c & 65.2  & 96.4 & \textbf{46.8} & \textbf{49.1} & 52.9 & 71.3 &  65.6 & \textbf{97.1} & 30.5 & \textbf{73.0} & 87.2 & \textbf{90.8} & 64.3 & 69.0 & 19.4 \\ \hline  
PyFu (Ours) & li+c & \textbf{67.8}& \textbf{96.5} & 39.6 &	49.0 &	 \textbf{75.3} & \textbf{82.2} & \textbf{67.2} & 96.3 & \textbf{39.6} & 71.4 & \textbf{87.7} & \textbf{90.8} & 60.6 & \textbf{70.5} & \textbf{22.7} \\ \hline
\end{tabular}
\vspace{-0.075cm}
\end{table*}
\section{CONCLUSION}
In this work, we presented a novel pyramid based fusion architecture \approach, which fuses lidar and camera features at multiple scales to improve 3D semantic segmentation. A Pyramid Fusion Backbone fuses the multi-scale features with a top-down and bottom-up strategy for an enhanced exploitation of multimodal information. It utilizes flexible fusion modules with interchangeable strategies. A Pyramid Fusion Head aggregates the pyramid features and refines them in a late fusion step. \approach\text{} outperforms other lidar range view and fusion approaches on two challenging outdoor datasets. The results underline the importance of sensor fusion in general and especially the benefits of the presented fusion architecture. It also emphasises, that the presented pyramid architecture exploits the sensor fusion best. To summarize, the proposed approach shows great potential in exploiting camera images to improve semantic segmentation of 3D point clouds.

\vspace{0.5cm}
\section*{ACKNOWLEDGMENT}
This work was supported by AUDI AG, Ingolstadt, Germany.
%

\newpage


\bibliographystyle{IEEEtran} 
\bibliography{IEEEabrv, reflist} 

\end{document}